\pdfoutput=1

\documentclass[11pt,a4paper]{article}

\usepackage[acceptedWithA]{tacl2018v2}

\usepackage{times}
\usepackage{latexsym}

\usepackage[T1]{fontenc}

\usepackage{microtype}

\usepackage{epsfig}
\usepackage{graphicx}
\usepackage{amsmath}
\usepackage{amssymb}
\usepackage{xspace}
\usepackage{amstext}
\usepackage{multirow}
\usepackage{tabularx}   
\usepackage{booktabs}   
\usepackage{arydshln}
\usepackage{subcaption}
\usepackage{colortbl}
\usepackage{xcolor}

\newcommand{\tblCross}[1]{\ensuremath{\textsc{CE}^{\textsc{#1}}}}
\newcommand{\tblEmb}[1]{\ensuremath{\textsc{BE}^{\textsc{#1}}}}
\newcommand{\tblConj}[1]{\ensuremath{\textsc{Joint+Coop}^{\textsc{#1}}}}
\newcommand{\tblCoop}[1]{\ensuremath{\textsc{Sep+Coop}^{\textsc{#1}}}}

\definecolor{Gray}{gray}{0.92}

\DeclareMathOperator*{\argmax}{arg\,max}

\title{Retrieve Fast, Rerank Smart: \\ Cooperative and Joint Approaches for Improved Cross-Modal Retrieval}

\author{\bf Gregor Geigle\thanks{{ } Both authors contributed equally to this work.}$^{\,\,\,1}$, Jonas Pfeiffer$^{*1}$, Nils Reimers$^{1}$,  \\
{\bf Ivan Vuli\'{c}$^{2}$, Iryna Gurevych$^{1}$ } \\
$^1$Ubiquitous Knowledge Processing Lab, 
  Technical University of Darmstadt \\
$^2$Language Technology Lab, University of Cambridge \hspace{0.5em} \\
{\url{www.ukp.tu-darmstadt.de}} \\
}

\date{}

\begin{document}

\maketitle

\begin{abstract}
\vspace{-0.3em}

Current state-of-the-art approaches to cross-modal retrieval process text and visual input jointly, relying on Transformer-based architectures with cross-attention mechanisms that attend over all words and objects in an image. While offering unmatched retrieval performance, such models: \textbf{1)} are typically pretrained from scratch and thus less scalable, \textbf{2)} suffer from huge retrieval latency and inefficiency issues, which makes them impractical in realistic applications. To address these crucial gaps towards both improved and efficient cross-modal retrieval, we propose a novel fine-tuning framework that turns any pretrained text-image multi-modal model into an efficient retrieval model. The framework is based on a cooperative retrieve-and-rerank approach which combines: \textbf{1)} twin networks (i.e., a bi-encoder) to separately encode all items of a corpus, enabling efficient initial retrieval, and \textbf{2)} a cross-encoder component for a more nuanced (i.e., smarter) ranking of the retrieved small set of items. We also propose to jointly fine-tune the two components with shared weights, yielding a more parameter-efficient model. Our experiments on a series of standard cross-modal retrieval benchmarks in monolingual, multilingual, and zero-shot setups, demonstrate improved accuracy and huge efficiency benefits over the state-of-the-art cross-encoders.\footnote{We release the code and model weights at \href{https://github.com/UKPLab/MMT-Retrieval}{github.com/UKPLab/MMT-Retrieval}.}

\end{abstract}

\vspace{-1.1em}
\section{Introduction}

Information-rich and efficient methods for dealing with large unstructured data in both computer vision and NLP are required to process and understand huge amounts of user-created content and beyond. In multi-modal contexts, such methods enable fundamental applications such as \textit{image retrieval}. A typical efficient \textit{bi-encoder}\footnote{Also frequently referred to as \textit{dual-encoder}.} approach encodes images and text \textit{separately} and then induces a shared high-dimensional multi-modal feature space. This enables cross-modal retrieval, where standard distance metrics identify the most similar examples for each query in the data collection via nearest-neighbor search \cite{Arya1998, KushilevitzOR00, LiuMGY04, AndoniI08, HajebiASZ11}.

\interfootnotelinepenalty=0

These bi-encoder approaches have already been shown to achieve reasonable performance in search and retrieval applications, both monolingually for English \cite{Nam2016, Faghri2017, zheng2017dual, wang2019-twobranch, shi2019knowledge} and in multilingual contexts \cite{gella2017pivoting:emnlp,kadar2018multilingual:conll,Kim2019,wehrmann2019language,Burns2020}. 
 
\begin{figure*}[t!]
    \centering

    \begin{subfigure}[!t]{0.155\linewidth}
        \centering
          \vspace{3em}
        \includegraphics[width=0.98\linewidth]{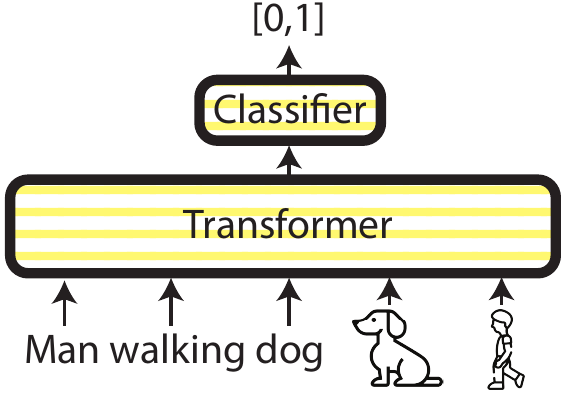}
      \vspace{1.5em}
        \caption{\tblCross{} Model}
        \label{fig:CE_ill}
    \end{subfigure}
    \hspace{1em}
        \begin{subfigure}[!t]{0.18\linewidth}
        \centering
         \vspace{2.1em}
        \includegraphics[width=0.975\linewidth]{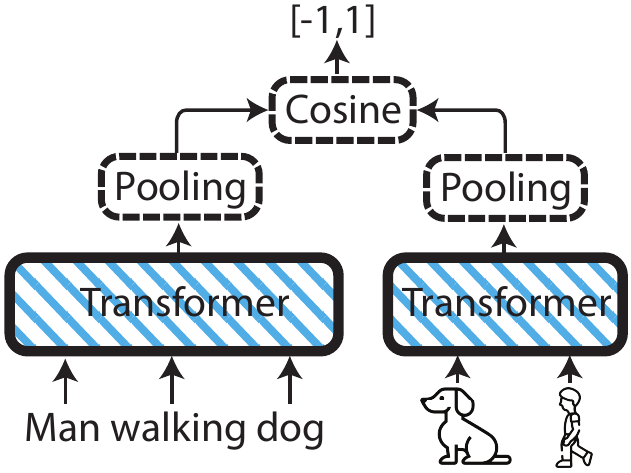}
        \vspace{1.4em}
        \caption{\tblEmb{} Model}
        \label{fig:emb_ill}
    \end{subfigure}
    \hspace{1em}
    \begin{subfigure}[!t]{0.18\linewidth}
        \centering
        \includegraphics[width=0.98\linewidth]{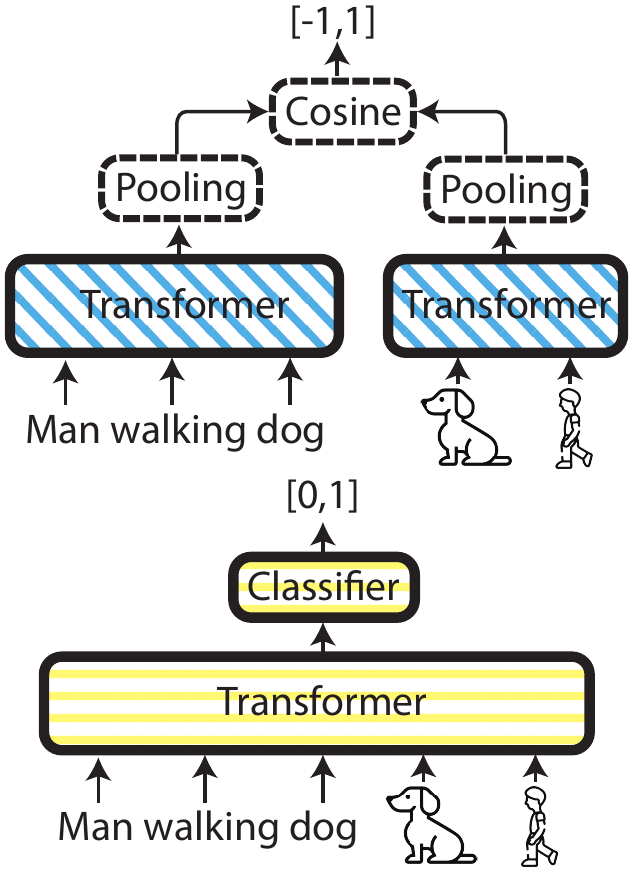}
        \caption{\tblCoop{}} 
        \label{fig:coop_ill}
    \end{subfigure}
    \hspace{1em}
        \begin{subfigure}[!t]{0.18\linewidth}
        \centering
        \includegraphics[width=0.98\linewidth]{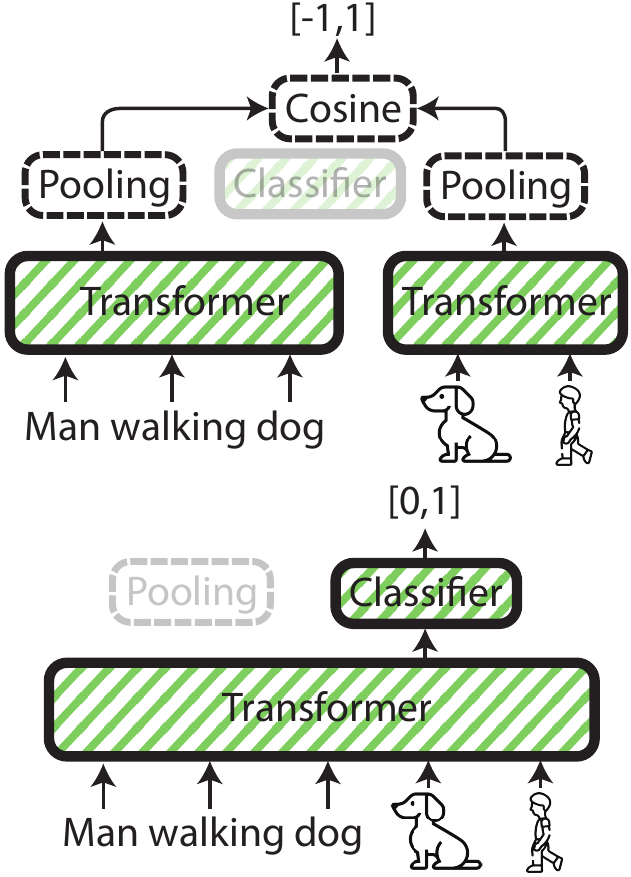}
        \caption{\tblConj{}} 
        \label{fig:conj_ill}
    \end{subfigure}
    \vspace{-1mm}
    \caption{Different architectures for image and text retrieval. Equal colors indicate shared weights. }
\label{fig:model_variants_ill}
\end{figure*}

However, they cannot match performance of more recent \textit{attention-based} methods. Here, a typical modus operandi is to apply a \textit{cross-attention} mechanism between examples from the two modalities to compute their similarity score, relying on Transformer-based neural architectures \cite{vaswani2017attention}. Such so-called multi-modal \textit{cross-encoders} (CE) \cite{tan2019lxmert, lu2019vilbert, Chen2019, Li2019, gan2020large, Li2020Oscar,Huang2020}  pass each text-image pair through the multi-modal encoder to compute their similarity, see Figure~\ref{fig:CE_ill}. 

While the results accomplished by the CE methods look impressive \cite{Li2020Oscar,bugliarello2020multimodal,Huang2020}, this comes at a prohibitive cost. In particular, they have extremely high search latency: processing a single text query with an image collection of 1M items may take up to 36 minutes using a single NVIDIA V100 GPU (see Table~\ref{tab:res:time-latency}). Due to this issue, they are evaluated only with extremely small benchmarks, i.e., the maximum size of typical image collections for image retrieval tasks is 5k images, and evaluation still lasts $\approx$50 hours (see Table~\ref{tab:res:time-ts}).\footnote{
Consequently, it would be impossible to evaluate these CE approaches on newer larger benchmarks: e.g., the (extrapolated) evaluation time on a benchmark spanning 100,000 images exceeds 2 years with a single GPU. 
} In sum, cross-encoders are impractical for deployment in realistic application scenarios, while the use of small benchmarks results in inflated and thus misleading evaluation performance.

In unimodal text-only setups, Transformer-based architectures have recently been integrated with bi-encoder (BE) methods \cite[\textit{inter alia}]{Guo:2018wmt,reimers2019sentence,humeau2019poly,Henderson:2019convert,Feng:2020labse}, yielding computationally more efficient sentence encoders. Instead of jointly encoding sentence pairs with cross-attention, a pretrained Transformer model (e.g., BERT \cite{Devlin2018}) is fine-tuned within a twin network with shared Transformer weights, as illustrated in Figure~\ref{fig:emb_ill}. In a nutshell, each sentence is passed through the encoder separately, and a loss function is defined on top of the two respective \textit{separately computed} encodings. However, despite their strong performance on sentence retrieval and similarity tasks \cite{reimers2019sentence,Litschko:2020ecir}, these encoders  cannot match the task performance of cross-encoders \cite{humeau2019poly}.

Motivated by these insights, in this work we aim to leverage \textit{the best of both worlds} towards improved and more efficient \textit{cross-modal search and retrieval}: \textbf{1)} efficiency and simplicity of \textsc{BE} approaches based on twin networks, as well as \textbf{2)} expressiveness and cutting-edge performance of \textsc{CE} methods. We first provide a systematic comparative analysis on the effectiveness and efficiency of Transformer-based multi-modal \textsc{BE} and \textsc{CE} methods across a range of image search evaluation benchmarks. We then propose two novel models which aim to blend the main strengths of \textsc{CE} and \textsc{BE}. The idea behind the first model variant, termed \textit{cooperative} (\tblCoop{}), is to \textit{retrieve and rerank} with two \textit{separate}, independently trained retrieval models: \textbf{1)} an initial \textit{top-$k$} list of potentially relevant items (i.e., texts or images) is retrieved by the more efficient \textsc{BE} model, and then \textbf{2)} this \textit{top-$k$} list is reranked \textit{``smartly''} by the more accurate \textsc{CE} model, as illustrated in Figure~\ref{fig:coop_ill}. Our second, \textit{joint} (\tblConj{}) model variant also operates in the same retrieve-and-rerank setup, but it now trains a multi-modal cross-encoder and a multi-modal \textsc{BE} model jointly with tied weights, as illustrated in Figure~\ref{fig:conj_ill}. The retrieve step, where efficiency is paramount, is again executed by the \textsc{BE} sub-model, and the precision-oriented rerank step is conducted via the \tblCross{} sub-model.

We propose a general framework for cross-modal search and retrieval, where \tblConj{} and \tblCoop{} models are independent of the chosen pretrained vision-language representation architectures. The experiments are thus based on a state-of-the-art vision-language architecture OSCAR \cite{Li2020Oscar}  
(experiments in English) and M3P \cite{Huang2020} (multilingual), and we demonstrate consistent improvements over the original OSCAR model on the standard benchmarks MSCOCO and Flick30k 
and improvements over the original M3P in multiple languages on the Multi30k dataset. We  empirically validate huge efficiency benefits of the proposed framework.

\vspace{1.6mm}
\noindent \textbf{Contributions.} \textbf{1)} We construct and systematically evaluate twin-networks combined with multi-modal Transformers (\tblEmb{}); they outperform all previous bi-encoder approaches, but lag behind their \tblCross{} counterparts.
\textbf{2)} We evaluate \tblEmb{} and \tblCross{} approaches within a cooperative retrieve-and-rerank approach; their combination outperforms the individual models, while offering substantial efficiency boosts compared to \tblCross{} methods. \textbf{3)} We propose a novel joint \tblCross{}-\tblEmb{} model (\tblConj{}), which is trained to simultaneously cross-encode and embed multi-modal input; it achieves the highest scores overall while maintaining retrieval efficiency. \textbf{4)} Finally, we propose a more realistic evaluation benchmark; we demonstrate harsh drops in overall cross-modal retrieval performance of all models in this more difficult scenario, calling for improved evaluation benchmarks and protocols in future work.

\section{Related Work}

Efficient approaches to cross-modal image-text retrieval relied on the induction of shared multi-modal visual-semantic embedding spaces (VSEs) \cite{Frome2013DiViSE, Faghri2017, shi2019knowledge, Mahajan2019JointWasserstein}. 
In a multilingual setup, all languages share the same embedding space along with the visual data \cite{Kim2019,wehrmann2019language,Burns2020}.
More recently, attention-based cross-encoder models, typically based on Transformer architectures \cite{vaswani2017attention} have considerably outperformed the VSE-based approaches. However, this comes at a severe cost of decreased retrieval efficiency and increased latency \cite{lee2018stacked, ijcai2019-pfan}. The current state-of-the-art multi-modal models jointly encode and cross-attend over text tokens and image features \cite[\textit{inter alia}]{lu2019vilbert, tan2019lxmert, Chen2019, Li2019, gan2020large, Li2020Oscar,bugliarello2020multimodal,Huang2020}. These CE methods leverage image captioning datasets such as MSCOCO \cite{lin2014microsoft} and Flick30k \cite{Plummer2015Flickr} and train a classification head that learns to identify whether or not an \textit{(image, caption) input pair} constitutes an aligned pair. Each image-text combination must be passed through the network, which scales quadratically with the number of examples.

To handle this quadratic increase, we use a cooperative retrieve-and-rerank approach.
While to the best of our knowledge this has not been proposed  for cross-modal settings, it has a long history in NLP, where
\citet{yates2021rankingoverview:sigir} date it back to the 1960s \cite{simmons1965questions}.
Until recently, bag-of-words methods (BoW; e.g., BM25) were commonly used for the first retrieval step.
For the second step, pretrained language models (LMs)
were fine-tuned to either rerank candidates \citep{nogueira2019rerank:arxiv,nogueira2019multistage:arxiv} or---for question-answering tasks---directly generated the answer span \citep{yang2019bertserini:naacl}.
More recent work on text-based retrieval and QA tasks has moved away from BoW methods towards learned (neural) models for the first retrieval step \citep{karpukhin2020dpr:emnlp,ding2020rocketqa,xiong2020approximate}.

Our work is inspired by recent \tblEmb{}-based approaches in unimodal text-only setups. Here, LMs 
are fine-tuned via twin-network architectures on auxiliary tasks such as semantic textual similarity \cite{reimers2019sentence,humeau2019poly}, paraphrasing \cite{Wieting:2019acl}, response retrieval \cite{Yang:2019repl,Henderson:2019acl,Henderson:2019convert,humeau2019poly}, or translation ranking \cite{Chidambaram:2019repl,Feng:2020labse}. This effectively turns the LMs into universal \textit{sentence encoders} which can then be used off-the-shelf for efficient text-based monolingual and cross-lingual retrieval \cite{Litschko:2020ecir}. In this work, we first extend this idea to multi-modal setups, and then show that our cooperative and joint approaches yields improved cross-modal retrieval models, maintaining retrieval efficiency.

Joint approaches like our \tblConj{} model that aim to align the retriever and reranker can be found in different forms:
\citet{boualili2020markedbert:sigir} ``mark'' exact matches from the bag-of-words retrieval for the reranker; \citet{yan2021rankingexpansion:aaai} share the parameters between a passage expander (which adds more relevant terms for a bag-of-words retriever) and the reranker;  \citet{hofstaetter2020marginmse:arxiv} distill knowledge from the reranker into the retriever model with soft labels generated by the teacher.
Specifically for questions-answering---where a two stage retriever-reader setup similar to the retrieve-and-rerank approach is common---research aims to synchronize the models through knowledge distillation from the reader to the retriever \citep{yang2020approximator:arxiv,izacard2021distilling:iclr} or by directly training both models end-to-end \citep{lee2019latentretrieval:acl,sachan2021endtoend:arxiv,sachan2021multidoc:arxiv}.
The challenge here is that the reader and the retriever are coupled---the reader requires candidates from the retriever that contain the solution. Our proposed reranker side-steps this problem as it uses no candidates from the retriever during training and only learns if a given input pair is (dis)similar. 
This way, we can train both components, the retriever and the reranker, side-by-side and align them by sharing their weights.

The work most closely related to ours includes contemporaneous models: 
ALBEF \citep{li2021alignbeforefuse:arxiv}, CLIP \cite{radford2021learning}, ALIGN \cite{Jia2021ALIGN}, 
and VisualSparta \cite{lu2021visualsparta}. 
ALBEF includes contrastive learning as one of its pretraining tasks but then uses a CE approach for downstream retrieval.
CLIP and ALIGN use similar contrastive learning strategies as we do, but are cast as full-fledged \textit{pretraining} architectures that learn from scratch and require magnitudes of more data than our approach. We show that it is possible to \textit{fine-tune} pretrained models with fewer data and offer a general framework, applicable to a spectrum of pretrained models. Further, unlike prior work, we demonstrate the benefits of combining \tblEmb{}-based (contrastive) learning with cross-encoders for improved and efficient retrieval.\footnote{As both CLIP \cite{radford2021learning} and ALIGN \cite{Jia2021ALIGN} disjoin the image and text components in their methods, cross-attention over the instances is not possible.} 
Finally, VisualSparta \cite{lu2021visualsparta} fine-tunes OSCAR, but at the level of token (text) and image-region embeddings. This enables the use of extremely fast lookup tables for efficient retrieval. However, this comes with a major disadvantage: the model disposes of wider context information.\footnote{E.g., considering a query ``two dogs and one cat'', the model is unable to match the numbers to the animals yielding likely worse retrieval results.} Our cooperative methods do leverage the finer-grained information at retrieval.

\section{Methodology}
\label{sec:methodology} 
The predominant Transformer-based multi-modal text-vision architecture is a single-stream encoder: it shares the majority of weights between the two modalities, including the multi-head cross-attention \cite{Chen2019, Li2019, gan2020large, Li2020Oscar,Huang2020}. 
The Transformer weights and text embeddings are typically initialized with weights of a pretrained LM (e.g., BERT \cite{Devlin2018} 
for English, XLM-R \cite{conneau2020xlmr} for multilingual models), where the corresponding vocabulary and tokenizer are utilized. Images are preprocessed via object detection models such as Faster R-CNN \cite{ren2015faster} to extract feature representations for regions of interest \cite{butd-anderson-2018}.
The image features are passed through an affine-transformation layer which learns to align the vision input with the pretrained Transformer. The position of the region of interest (or in some models also the region's width and height) is used to generate positional embeddings.  By combining these two representations, each object region is passed into the Transformer separately.
The cross-attention mechanism of the Transformer attends over all text and image inputs at every layer, thus learning a joint representation of both modalities.

Similar to masked language modeling (MLM) in the text domain, 
multi-modal Transformer models are trained with self-supervised objectives.  
For pretraining, image-caption datasets (i.e.,  MSCOCO \cite{lin2014microsoft}, Flickr30k \cite{Plummer2015Flickr}, Conceptual Captions (CC) \cite{sharma-etal-2018-conceptual}, and SBU \cite{Ordonez:2011:im2text}) are utilized. The pretrained multi-modal model is subsequently fine-tuned with multi-modal downstream task data. 

We focus on different \textit{fine-tuning strategies} of the pretrained models for the downstream task of image-text retrieval. We illustrate these approaches in Figure \ref{fig:model_variants_ill} and describe them in what follows.

\subsection{Cross-Encoders}
\label{sec:cross_enc}
For image and text retrieval tasks, the prevailing approach with pretrained multi-modal Transformer models is to cross-encode each image-text combination (see Figure~\ref{fig:CE_ill}).

\vspace{1.6mm}
\noindent \textbf{Training.}
A pretrained model receives as input positive and negative pairs of images and captions. Negative pairs are also sampled from the training dataset (e.g., MSCOCO, Flickr30k). A binary classification head is placed on top of the Transformer model, where the contextualized embedding of the \texttt{[CLS]} token is passed into the classification head. The weights of the classifier together with the Transformer, word embeddings, and image feature transformation matrices are fully fine-tuned using a binary cross-entropy (BCE) loss:

\vspace{-2mm}
{\footnotesize
\begin{align}
    \mathcal{L}_{\tblCross{}}(i,c) =  -\Big(y \log \text{p}(i,c) + (1-y) \log (1-\text{p}(i,c)) \Big). \notag
    \label{eq:cross_entropy_loss}
\end{align}
 }%
$\text{p}(i,c)$ indicates the probability of the input combination of image $i$ and caption $c$ to have the positive label (i.e., whether it is the correct image-caption combination); $y=1$ if $(i,c)$ is a positive pair and $y=0$ if either the image or text has been replaced (i.e., a negative pair).\footnote{Some cross-encoders such as UNITER \cite{Chen2019} and VL-BERT \cite{su2019vl} rely on another triplet loss function \cite{Chechik:2010jmlr}; however, OSCAR \cite{Li2020Oscar} reports improved performance with BCE.}

\vspace{1.6mm}
\noindent \textbf{Retrieval.}
At retrieval, all $(i,c)$ combinations need to be processed, and are ranked by the probability $p(i,c)$. For instance, given a text query $c$, retrieving the single most relevant image $i$ from an image collection $I$ proceeds as follows:
\begin{equation}
\small
    \argmax(\text{p}(i,c), \, \forall i \in I)
    \label{eq:ceretrieval}
\end{equation}
Despite its typically high performance, this approach comes at high computational costs as each target instance needs to be passed through the entire network along with the query to obtain the score $p(i,c)$; that is, the approach does not leverage any pre-computed representations during retrieval.

\subsection{Bi-Encoders}
\label{sec:embedding}

\noindent \textbf{Training.}
Each image and text caption is passed separately through the pretrained Transformer model (Figure~\ref{fig:emb_ill}). The contextualized representations are mean-pooled to represent the embedding of the respective image $\mathbf{i}$ and text caption $\mathbf{c}$.\footnote{Following \citet{reimers2019sentence} we opt for mean pooling as the final ``aggregated'' embedding; it outperformed another standard variant, which uses the \texttt{[CLS]} token, in our preliminary experiments.} 
The objective of the twin network is to place positive training instances $(i,c)$ closely in the shared multi-modal space, while unrelated instances should be placed farther apart. This is formulated through a standard triplet loss function. It leverages $(i,c,c')$ and $(i,i',c)$ triplets, where $(i,c)$ are positive image-caption pairs from the training corpus, while $c'$ and $i'$ are negative examples sampled from the same corpus such that image-caption pairs/instances $(i,c')$ and $(i',c)$ do not occur in the corpus. The triplet loss is then: 
\begin{multline}
 \small        \mathcal{L}_{\tblEmb{}}(i,c) = [\text{cos}(\mathbf{i},\mathbf{c}') - \text{cos}(\mathbf{i},\mathbf{c}) + \alpha]^+  \\
  \small   + [\text{cos}(\mathbf{i}',\mathbf{c}) - \text{cos}(\mathbf{i},\mathbf{c}) + \alpha]^+ 
    \label{eq:triplet_loss}
\end{multline}
\noindent where $[\cdot]^+=\mathrm{max}(0,\cdot)$, $\alpha$ defines a margin, and $\mathbf{i}'$ and $\mathbf{c}'$ are embeddings of respective image and caption negatives.

\vspace{1.6mm}
\noindent \textbf{Sampling Negative Examples.}
 Negative examples may have a profound impact on training and performance, and it has been shown that selecting hard negative examples typically yields improved performance 
 \cite{Faghri2017}. However, detecting such hard negatives is only possible with \tblEmb{}-based approaches, as cross-encoding all instances is computationally infeasible. We rely on the \textit{In-Batch Hard Negatives (BHN)} method \cite{HermansBeyer2017Arxiv}, a computationally efficient sampling of hard negative examples. In a nutshell, BHN randomly samples a set of $N$ negative examples from the training corpus and then ranks them according to their distance to all positive examples;  for each positive example, the closest negative example is selected as the \textit{hardest} negative example. By scaling up $N$, the probability of sampling truly hard negatives increases.

\vspace{1.6mm}
\noindent \textbf{Retrieval.}
The \tblEmb{} approach enables \textit{pre-encoding} of all items for efficient retrieval look-up.\footnote{Note that pre-computing the embedding does come with increased storage and memory demands; e.g., with a base Transformer architecture this requires an additional $\approx3$kB of memory for each embedding. A corpus of 1M images would amount to $\approx3$GB of required storage.}
For instance, a text query $q$ is encoded with the bi-encoder and the most similar pre-encoded instance from an image collection $I$ is retrieved: 
    $\argmax_{i \in I}\text{cos}(\mathbf{i}, \mathbf{q})$.

This approach can  scale to even billions of images \cite{JDH17}, but it cannot be guaranteed that the important idiosyncratic information necessary to distinguish truly relevant from related examples is sufficiently encoded in the embedding. Further, the approach might not generalize well in low-resource scenarios as the model is not required to learn finer-grained parts of the input if they are never demanded by the training data. 

\subsection{Separate Training, Cooperative Retrieval}
\label{sec:cooperative}
We  combine the benefits of the two model types (\tblCross{} and \tblEmb{}) within a \textit{cooperative} retrieval approach (\tblCoop{}), as illustrated in Figure~\ref{fig:coop_ill}.

\vspace{1.6mm}
\noindent \textbf{Training and Retrieval.} 
Two models, one \tblCross{} (\S\ref{sec:cross_enc}) and one \tblEmb{} (\S\ref{sec:embedding}), are trained independently. Following that, the retrieval step is split into two stages. First, the efficient \tblEmb{} model is used to retrieve the \textit{top-$k$} relevant items from the entire large collection, yielding a much smaller collection $I_k$: 
    $I_k = \text{top}_k(\{\text{cos}(\mathbf{i}, \mathbf{q}): \, \forall i \in I\})$, 
where $\text{top}_k(\cdot)$ retrieves a set of the top-$k$ most similar instances. Second, we rerank the instances from $I_k$ with the more precise but computationally more expensive \tblCross{} model: 
    $\argmax_{i \in I'}\text{p}(i,c)$. 
This cooperative approach thus combines the benefits of both approaches and is able to efficiently retrieve instances.\footnote{Retrieval time for  1M images: 94ms (GPU), 13s (CPU).} However, given that this approach requires two models to be stored in memory, it is less parameter-efficient than the previous methods.

\subsection{Joint Training, Cooperative Retrieval}
\label{sec:conjoint}

\noindent \textbf{Training and Retrieval.}
Instead of relying on two fully separated models, we propose to train a single joint model, able to both \textit{cross-encode} and \textit{embed} (i.e., `\textit{bi-encode}'), see Figure \ref{fig:conj_ill}. The joint model with shared parameters trains by alternating between the respective sub-models and their input types. When cross-encoding, a dedicated prediction head is trained using BCE loss (\S\ref{sec:cross_enc}). In order to train the \tblEmb{}-based sub-model, we again rely on a twin architecture with a triplet loss from Eq.~\eqref{eq:triplet_loss}.

Retrieval proceeds with the same two-step retrieve-and-rerank procedure from \S\ref{sec:cooperative}. We first obtain the set $I_k$ with the much cheaper \tblEmb{}-based submodel, and then rerank its items with the CE submodel. We combine the best traits of \tblCross{} and \tblEmb{}, while maintaining parameter efficiency. Using both learning objectives at training, the joint model is forced to observe the input from different viewpoints, thus improving its generalization capability while offering parameter efficiency.

\section{Experimental Setup}
\label{sec:experimental}
Our fine-tuning framework from \S\ref{sec:methodology} can be applied to any pretrained multi-modal Transformer. In all the experiments, we opt for state-of-the-art pretrained multi-modal models for monolingual (English) and multilingual contexts: OSCAR \cite{Li2020Oscar} and M3P \cite{Huang2020}, respectively.

\vspace{1.3mm}
\noindent \textit{OSCAR} is a single-stream multi-modal Transformer \cite{bugliarello2020multimodal}, with its weights initialized with those of the pretrained BERT Base model, and then subsequently fine-tuned on multi-modal data (see \S\ref{sec:methodology}). Unlike prior work, OSCAR additionally uses object labels of detected regions: those labels serve as anchors for visual grounding, with large improvements achieved over its prior work. \textit{M3P} is a single-stream multilingual multi-modal Transformer. Its weights are initialized with those of pretrained XLM-R Base and then fine-tuned on multi-modal data (see \S\ref{sec:methodology}) as well as multilingual text-only data. 

\vspace{1.6mm}
\noindent \textbf{Training and Test Data.} We primarily experiment with the English image-text retrieval benchmarks MSCOCO and Flick30k. They respectively comprise 123k and 31.8k images, with  5 captions describing each image. MSCOCO provides two test benchmarks of sizes 1k and 5k, where the smaller set is a subset of the 5k test set. The standard Flickr30k test set consists of 1k images. In addition, we use the development set of Conceptual Captions (CC) \cite{sharma-etal-2018-conceptual} for zero-shot evaluation, and also to construct a larger and more difficult test set (see later in \S\ref{sec:further}). The original CC dev set contained 15.8k images, but currently, only 14k images are still available online.

For multilingual experiments, we use the standard Multi30k dataset \cite{elliott-etal-2016-multi30k,elliott-etal-2017-findings,barrault-etal-2018-findings}, which extends Flickr30k with 5 German and one French and Czech caption per image. Its test set also comprises 1k images.

The evaluation metric is the standard \textit{Recall-at-M (R@M)}: it reports the proportion of queries for which the relevant target item is present within the top-$M$ retrieved items.

\vspace{1.6mm}
\noindent \textbf{Training Setup and Hyperparameters.} Our setup largely follows \citet{Li2020Oscar} and \citet{Huang2020} unless noted otherwise.\footnote{Unlike \citet{Li2020Oscar} we do not use object tags as additional input, as preliminary experiments suggested no improvement with object tags.} We experiment with learning rates $[5e-5, 2e-5]$, and with the number of update steps between 25k and 125k. One batch contains 128 positive pairs plus 128 negative pairs with $\mathcal{L}_{\tblCross{}}$. We use the AdamW optimizer \cite{Loschilov:2018iclr} with a linear learning rate  decay without warmup, and a weight decay of 0.05. We take model checkpoints every 5k steps and select the checkpoint with the best development set performance.

\subsection{Baselines and Model Variants}
\label{sec:variants}

\noindent \textbf{\tblCross{}.} 
Our main baselines are   OSCAR and M3P models used in the standard \tblCross{} setting, described in \S\ref{sec:cross_enc}. We fully fine-tune the Transformer weights along with a randomly initialized classification head.\footnote{Training for 100k steps and a learning rate of $2e-5$ (OSCAR) or $5e-5$ (M3P) performed best.} At retrieval, we cross-encode each text-image combination and rank them according to the corresponding probability, see Eq.~\eqref{eq:ceretrieval}.

\vspace{1.3mm} 
\noindent \textbf{\tblEmb{}{}.} 
We rely on  BHN negative sampling, 
finding that training for 30k steps, with a learning rate of $5e-5$, and with a margin $\alpha=0.1$ works best.\footnote{We also experimented with \textit{Approximate-nearest-neighbor Negative Contrastive Estimation (ANCE)} \cite{xiong2020approximate}; however, it did not yield performance benefits.  
}

\vspace{1.3mm}
\noindent \textbf{\textsc{Sep+Coop}.} 
For the cooperative method without joint training (\S\ref{sec:cooperative}),  
we retrieve the top-20 instances with \tblEmb{}{} 
and rerank them via \tblCross{}.\footnote{We provide an ablation study of different $k$ values in \S \ref{sec:further}. We have also experimented with training a \tblCross{} 
model using hard negative samples from a pretrained \tblEmb{}{} model
. However, the \tblCross{} 
model is able to easily overfit on those negative examples, resulting in inferior performance.}

\vspace{1.3mm}
\noindent \textbf{\textsc{Joint+Coop}.}  
We alternate between the two objective functions while training the joint model (see \S\ref{sec:conjoint}). We find that training for 60k update steps with a learning rate of $2e-5$ (OSCAR) or $5e-5$ (M3P) works best, the rest of the hyperparameters are the same as with separately trained models. For retrieval, we again set $k=20$. To demonstrate the benefits of cooperative retrieval, we also evaluate two non-cooperative variants originating from the joint model: \textbf{\textsc{Joint+CE}} uses the \tblCross{} sub-model for a single-step CE-style retrieval, while \textbf{\textsc{Joint+BE}} operates in the fully \tblEmb{} retrieval setup.

\vspace{1.4mm}
\noindent The underlying pretrained Transformer is denoted with a superscript: e.g., \textsc{Joint+Coop}$^\text{OSCAR}$ denotes that: 1) pretrained OSCAR is 2) fine-tuned with the joint variant from \S\ref{sec:conjoint}, and 3) then used in the cooperative retrieval setup.

\begin{table*}[!t]
    \centering
    \def\arraystretch{0.93}
\resizebox{0.99\textwidth }{!}{
    \begin{tabular}{cl rrr rrr rrr rrr}
    \toprule
 \bf Group    &   \bf Model &  \multicolumn{3}{c}{\bf Image Retrieval} & \multicolumn{3}{c}{\bf Text Retrieval}  & \multicolumn{3}{c}{\bf Image Retrieval} & \multicolumn{3}{c}{\bf Text Retrieval}\\
  &      & R@1 & R@5 & R@10 & R@1 & R@5 & R@10 & R@1 & R@5 & R@10 & R@1 & R@5 & R@10\\
\cmidrule(lr){3-5} \cmidrule(lr){6-8} \cmidrule(lr){9-11} \cmidrule(lr){12-14}
& &  \multicolumn{6}{c}{MSCOCO (5k)} & \multicolumn{6}{c}{Flickr30k (1k)} \\
\cmidrule(lr){3-5} \cmidrule(lr){6-8} \cmidrule(lr){9-11} \cmidrule(lr){12-14}
\multirow{4}{*}{\rotatebox[origin=c]{0}{\begin{tabular}[c]{@{}c@{}} G1. Pre-Transformer \end{tabular}}} & VSE++ \cite{Faghri2017} & \bf 43.9 & 59.4 & 72.4& 41.3 & 71.1 & 81.2 & 39.6  & 70.1  & 79.5 & 52.9  & 80.5  & 87.2 \\
  & SCAN \cite{lee2018stacked} &  38.6 & \bf 69.3 & 80.4& 50.4 & 82.2 & 90.0& 48.6  & 77.7  & 85.2 & 67.9  & 90.3  & \bf 95.8   \\
  & PFAN \cite{ijcai2019-pfan} &  \multicolumn{1}{c}{\, ---} & \multicolumn{1}{c}{\, ---} & \multicolumn{1}{c}{\, ---} & \multicolumn{1}{c}{\, ---} & \multicolumn{1}{c}{\, ---} & \multicolumn{1}{c}{\, ---}  & \bf 50.4 & \bf 78.7 & \bf 86.1& 70.0 & \bf  91.8 & 95.0 \\
  & SCG \cite{shi2019knowledge} &  39.2 & 68.0 & \bf 81.3& \bf 56.6 & \bf 84.5 & \bf 92.0 & 49.3  & 76.4  & 85.6 & \bf 71.8  & 90.8  & 94.8  \\
\cmidrule(lr){3-5} \cmidrule(lr){6-8} \cmidrule(lr){9-11} \cmidrule(lr){12-14}
\multirow{5}{*}{\rotatebox[origin=c]{0}{\begin{tabular}[c]{@{}c@{}}   G2. Cross-Encoders\\ \textit{(Inefficient} \\ \textit{for  retrieval)} \end{tabular}}}  & \tblCross{}$^\text{UNITER}$ \cite{Chen2019} &  48.4 & 76.7 & 85.0& 63.3 & 87.0 & 93.1 & 72.5  & 92.4  & 96.1 & 85.9  & 97.1  & 98.8  \\
  & \tblCross{}$^\text{Unicoder-VL}$ \cite{Li2019} &  46.7 & 76.0 & 85.3& 62.3 & 87.1 & 92.8 & 71.5  & 90.9  & 94.9 & 86.2  & 96.3  & 99.0 \\ 
  & \tblCross{}$^\text{VILLA}$ \cite{gan2020large} &  \multicolumn{1}{c}{\, ---} & \multicolumn{1}{c}{\, ---} & \multicolumn{1}{c}{\, ---} & \multicolumn{1}{c}{\, ---} & \multicolumn{1}{c}{\, ---} & \multicolumn{1}{c}{\, ---}  & 74.7 & 92.9 & 95.8& 86.6 & 97.9 & \underline{\bf 99.2} \\

  & \tblCross{}$^\text{OSCAR}\dagger$ \cite{Li2020Oscar}& \bf  54.0 & \bf  80.8 & \bf 88.5& \bf  70.0 & \bf 91.1 & \underline{\bf 95.5}   & \multicolumn{1}{c}{\, ---} & \multicolumn{1}{c}{\, ---} & \multicolumn{1}{c}{\, ---} & \multicolumn{1}{c}{\, ---} & \multicolumn{1}{c}{\, ---} & \multicolumn{1}{c}{\, ---} \\
  & \tblCross{}$^\text{OSCAR}\ddagger$ &   52.6 & 80.0 & 88.1 & 69.3 & 90.7 & 95.3 & \bf  75.9 & \bf  93.3 & \underline{\bf 96.6} & \bf  88.5 & \underline{\bf 98.5} & \underline{\bf 99.2}\\
\cmidrule(lr){3-5} \cmidrule(lr){6-8} \cmidrule(lr){9-11} \cmidrule(lr){12-14}
 \multirow{4}{*}{\rotatebox[origin=c]{0}{\begin{tabular}[c]{@{}c@{}}   G3. Bi-Encoders  \\  \textit{(Efficient} \\  \textit{for  retrieval)} \end{tabular}}}   & VisualSparta \cite{lu2021visualsparta} & 44.4 & 72.8 & 82.4 &\multicolumn{1}{c}{\, ---}&\multicolumn{1}{c}{\, ---}&\multicolumn{1}{c}{\, ---} & 57.4 & 82.0 & 88.1 &\multicolumn{1}{c}{\, ---}&\multicolumn{1}{c}{\, ---}&\multicolumn{1}{c}{\, ---}\\ 
  & \tblEmb{}$^\text{OSCAR}$ &  52.2 & 80.2 & 88.0 & 66.9 & 90.1 & 95.0 & 72.0 & 91.0 & 94.7 & 84.7 & 97.1 & 98.7 \\
  & \textsc{Sep+Coop}$^\text{OSCAR}$ &  52.8 & 80.5 & 88.5 & 70.2 & \underline{\bf  91.6} & 95.0 & 76.0 & 93.0 & 95.0 & 88.7 & \bf  98.3 & \underline{\bf 99.2} \\
  & \textsc{Joint+Coop}$^\text{OSCAR}$ & \underline{ \bf 54.7} & \underline{\bf 81.3} & \underline{\bf 88.9} & \underline{\bf 70.8} & 91.0 & \bf  95.2 & \underline{\bf 76.4} & \underline{\bf 93.6} & \bf   96.2 & \underline{\bf 89.4} & 97.7 & 99.0 \\
\cmidrule(lr){3-5} \cmidrule(lr){6-8} \cmidrule(lr){9-11} \cmidrule(lr){12-14}
  & \textsc{Joint+CE}$^\text{OSCAR}$ &  54.6 & 81.1 & 88.8 & 70.6 & 91.0 & 95.1 & 76.5 & 93.4 & 96.3 & 89.0 & 97.9 & 99.1 \\
  & \textsc{Joint+BE}$^\text{OSCAR}$ &  52.5 & 80.0 & 88.0 & 66.7 & 90.0 & 95.0 & 71.6 & 91.5 & 95.0 & 86.3 & 96.8 & 98.6\\
        \bottomrule
    \end{tabular}
    }
    \caption{Results on MSCOCO and Flickr30k (monolingual setups). The group G1 presents results from the literature with Pre-Transformer (PT) approaches. G2 denotes the results of recent cross-encoders with Transformers (\tblCross{}$^*$; \S\ref{sec:cross_enc}). 
  $\dagger$ indicates the results taken directly from the literature \cite{Li2020Oscar}, while $\ddagger$ indicates our own results achieved with the model weights. G3 covers efficient retrieval methods that either retrieve images based only on distance metrics (\tblEmb{}, \S\ref{sec:embedding}), or rely on the \tblCoop{} approach (see \S\ref{sec:cooperative} and \S\ref{sec:conjoint}). The last two lines present the results of the joint model without the cooperative retrieval step (see \S\ref{sec:variants}). Highest results per each group in \textbf{bold}, highest overall results are \underline{underlined.}  
    }
    \label{tab:main_results}
    \vspace{1em}
\end{table*}

\begin{table}[!t]
    \footnotesize
    \centering
    \def\arraystretch{0.87}
    \resizebox{0.99\columnwidth}{!}{
    \begin{tabular}{cl rrrrr}
    \toprule
          \bf Type    &  \bf Model & \bf en & \bf de & \bf fr & \bf cs & \bf mean\\
         \cmidrule(lr){3-6}  \cmidrule(lr){7-7}
\multirow{3}{*}{\rotatebox[origin=l]{0}{\begin{tabular}[l]{@{}c@{}} G1. PT \end{tabular}}} & MULE  & 70.3 &64.1 & 62.3 & 57.7 & 63.6\\
& S-LIWE   & 76.3 & 72.1 & 63.4 & 59.4 & 67.8\\
& SMALR   & 74.5 & 69.8 & 65.9 & 64.8 & 68.8\\
\cmidrule(lr){3-6}  \cmidrule(lr){7-7}
\multirow{2}{*}{\rotatebox[origin=l]{0}{\begin{tabular}[l]{@{}c@{}}   G2.  CE \end{tabular}}} & \tblCross{}$^\text{M3P}\dagger$   & \bf 86.7 & \bf 82.2 & 73.5 & 70.2 & 78.2\\
& \tblCross{}$^\text{M3P}\ddagger$  & 83.7 & 79.4 & 76.5 & 74.6 & 78.6\\
\cmidrule(lr){3-6}  \cmidrule(lr){7-7}
 \multirow{3}{*}{\rotatebox[origin=l]{0}{\begin{tabular}[l]{@{}c@{}}   G3. BE \end{tabular}}} & \tblEmb{}$^\text{M3P}$   & 82.8 & 78.0 & 75.1 & 73.6  & 77.4\\
& \textsc{Sep+Coop}$^\text{M3P}$  & 84.8 & 80.5 & \bf 77.5 & \bf 75.6  & \bf 79.6\\
& \textsc{Joint+Coop}$^\text{M3P}$ & 83.0 & 79.2 & 75.9 & 74.0   & 78.0\\
        \bottomrule
    \end{tabular}
    }
    \caption{Results on Multi30k (multilingual setups). Following prior work \cite{Huang2020}, we report \textit{mean Recall (mR)} scores: mR computes an average score of Recall@1, Recall@5 and Recall@10 on image-to-text retrieval and text-to-image retrieval tasks. All methods in the comparison use text data from all four languages. We divide the models into groups G1-G3 as in Table~\ref{tab:main_results}. $\dagger$ indicates results taken directly from the literature \cite{Huang2020} and $\ddagger$ indicates our own results. MULE \cite{Kim2019}; S-LIWE \cite{wehrmann2019language}; SMALR \cite{Burns2020}; \tblCross{}$^\text{M3P}\dagger$ \cite{Huang2020}.
    } 
    \label{tab:res:multilingual-all}
\end{table}
\begin{figure*}[!t]
    \centering
    Caption: \textit{A skier is skiing down the snow wearing a white shirt and black shorts.}\\
    \begin{subfigure}[!t]{0.16\linewidth}
        \centering
        \includegraphics[width=0.96\linewidth]{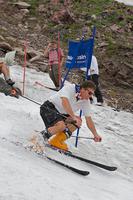}
        \caption{Target Image}
        \label{fig:CE_ill1}
    \end{subfigure}
    \begin{subfigure}[!t]{0.725\linewidth}
        \begin{subfigure}[!t]{\linewidth}
        \centering
        \includegraphics[width=0.95\linewidth]{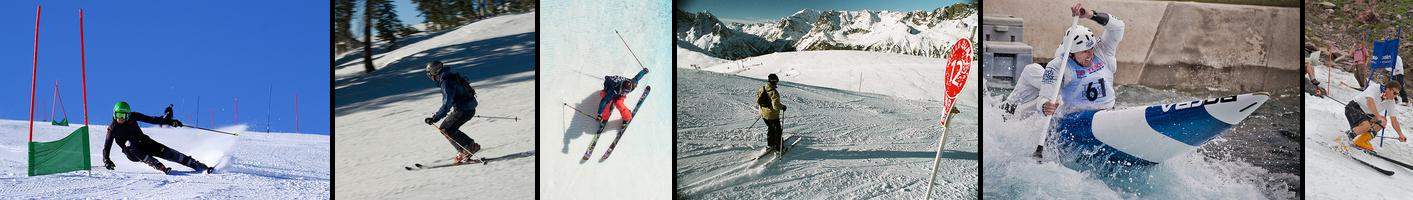}
        \caption{Retrieved Top-6 with \textsc{Joint+BE}$^\text{OSCAR}$}
        \label{fig:emb_ill1}
    \end{subfigure}
    \begin{subfigure}[!t]{\linewidth}
        \centering
        \includegraphics[width=0.95\linewidth]{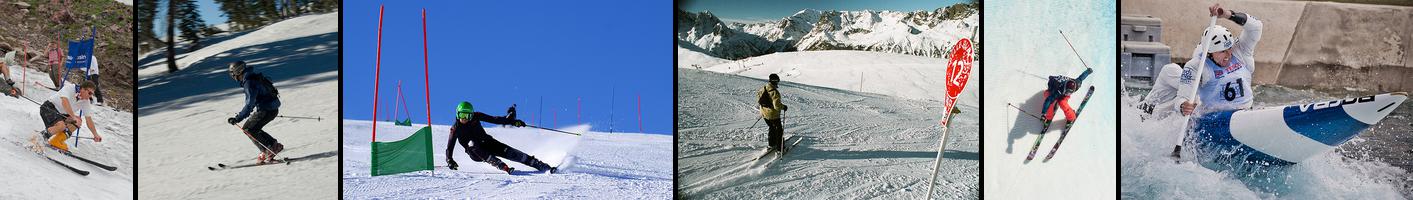}
        \caption{Reranked Top-6  with \textsc{Joint+CE}$^\text{OSCAR}$}
        \label{fig:coop_ill1}
    \end{subfigure}
    \end{subfigure}

    \caption{
    We efficiently retrieve the top instances with the  \textsc{Joint+BE}$^\text{OSCAR}$ submodel to identify the (globally) most relevant target instances. The more precise, but less efficient \textsc{Joint+CE}$^\text{OSCAR}$ submodel  then disentangles the specific intricacies of the  images. Ranking proceeds from left to right. 
    }
\label{fig:example-rerank}
\vspace{-1.5mm}
\end{figure*}

\section{Results and Discussion}
The main results on English-only monolingual datasets Flickr30k and MSCOCO are summarized in Table~\ref{tab:main_results}, while the scores on multilingual Multi30k are provided in Table~\ref{tab:res:multilingual-all}. 

As expected, all Transformer-based approaches (groups G2 and G3) substantially outperform the pre-Transformer (PT) models (G1). While this has already been established in prior work for \tblCross{} methods, our findings confirm that the same holds also for the efficient \tblEmb{} approach. This validates the effectiveness of Transformer architectures pretrained on large corpora for the retrieval task. R@1 scores with \tblEmb{} lag slightly behind the \tblCross{} scores, but the respective R@10 scores are mostly on-par. This suggests that the \tblEmb{} approach is ``coarser-grained'', and mostly relies on ``global'' interactions between the modalities. We investigate this conjecture further in \S\ref{sec:further}. 

This is also illustrated by an example in Figure~\ref{fig:example-rerank}. 
When dealing with related target items, \tblCross{}'s cross-attention mechanism is able to explicitly attend over each token and image region, capturing additional (non-global) information relevant to the query.  
 While the high-level ``global'' concept of a \textit{skiing person} is present in (almost) every example, the additional important information related to \textit{what the person is wearing} is not adequately represented  in the embeddings. Therefore, the \tblEmb{} (sub)model does not rank this instance at the top position. The \tblCross{} (sub)model then directly compares the instances, 
identifying that clothing is  important  and reranks the target examples accordingly.

Most importantly, the relative comparison of R@1 versus R@10 scores empirically hints at the necessity of the retrieve-and-rerank cooperative approach: the \tblEmb{} approach efficiently retrieves 20 relevant examples, but the increased expressiveness of \tblCross{} is required to refine the initially retrieved list. Moreover, the results in the cooperative setup even without joint training (\textsc{Sep+Coop}$^{\text{OSCAR}}$ and \textsc{Sep+Coop}$^{\text{M3P}}$) demonstrate that the two models support each other: slight improvements are observed over the pure \tblCross{}, while offering massive efficiency boosts over \tblCross{}. Our speculation is that the \tblEmb{} model filters out false positives, which in turn makes the \tblCross{} model more robust. 

The results of the \textsc{Joint+Coop} variant indicate that it is indeed possible to maintain retrieval efficiency with improved parameter efficiency: this approach performs on-par or even slightly outperforms the standard state-of-the-art \tblCross{} models.  The results verify that the two objective functions do not interfere with each other and that a single model is able to both embed and cross-encode. We note that the \textsc{Joint+Coop} variant offers the best trade-off between parameter and retrieval efficiency, achieving the peak scores on the monolingual MSCOCO and Flickr30k benchmarks, and very competitive results on the multilingual Multi30k benchmark.

\section{Further Analysis}
\label{sec:further}
We now discuss a series of additional experiments which further profile and analyze the proposed multi-modal retrieval approaches, focusing especially on the multiple efficiency aspects related to fine-tuning and retrieval stages.

\begin{table}[]
    \centering
     \def\arraystretch{0.93}
     {\footnotesize
    \begin{tabularx}{0.97\linewidth}{l XX XX}
    \toprule
         \bf Model & \multicolumn{2}{c}{\bf NVIDIA V100} & \multicolumn{2}{c}{\bf CPU}\\
&  \bf 50k & \bf 1M & \bf 50k & \bf 1M\\
\cmidrule(lr){2-3} \cmidrule(lr){4-5}
\tblEmb{} & 16ms & 37ms & 0.2s & 1.6s \\
\textsc{Sep/Joint+Coop} & 74ms & 94ms & 6s & 13s \\
\tblCross{} & 2min  & 36min & 2.4h & 47h \\
        \bottomrule
    \end{tabularx}
    }%
    \caption{Retrieval latency for one query with an image collection of 50k or 1M images (with pre-encoded images) using a single GPU/ CPU. Batch size for cross-encoding of the query with the images is 512. CPU is an Intel Xeon Gold 6154.}
    \label{tab:res:time-latency}
\end{table}

\begin{table}[!t]
    \centering
     \def\arraystretch{0.93}
     {\footnotesize
    \begin{tabularx}{0.97\linewidth}{l XXX}
    \toprule
         \bf Model & \bf 1k & \bf 5k & \bf 100k\\
\cmidrule(lr){2-4}
\tblEmb{} & 5s & 30s & 7min\\
\textsc{Sep/Joint+Coop} & 5min & 25min & 8.5h\\
\tblCross{} & 2h & 50h & 2.3a*\\
        \bottomrule
    \end{tabularx}
    }%
    \caption{Evaluation time for the  MSCOCO test sets of 1k, 5k, and 100k images on an NVIDIA V100 with batch size 512. The time includes bi-encoding images \& text, i.e., the embeddings are not pre-computed.  * denotes extrapolated values.}
    \label{tab:res:time-ts}
\end{table}

\vspace{1.6mm}
\noindent \textbf{Retrieval Efficiency.}
We empirically validate the time efficiency of our cooperative approaches for retrieval in an image search scenario (Tables~\ref{tab:res:time-latency}) and for evaluation on huge datasets (Tables~\ref{tab:res:time-ts}).
To allow for a fair comparison between the approaches, we implement the entire retrieval pipeline---from model to nearest-neighbor search---in Pytorch without additional optimization such as multi-processing or optimized nearest-neighbor search libraries like FAISS \citep{JDH17}. 

Our measurements confirm the efficiency of BEs in comparison to CEs.
The cooperative approaches, which only have to cross-encode a constant number of items invariant of the collection size, are close in retrieval latency to \tblEmb{} for image search and remain feasible even for large datasets.

\vspace{1.9mm} 
\noindent \textbf{Larger Benchmarks.} The results in Table~\ref{tab:main_results} indicate that current top-performance models achieve very high scores in absolute terms on the standard retrieval benchmarks. However, this is partially due to too small image collections with only a few thousand instances; one undesired effect is that it becomes increasingly difficult to identify significant differences between model performances. Unfortunately, the inefficiency of \tblCross{} models
, as empirically validated in Tables~\ref{tab:res:time-latency}-\ref{tab:res:time-ts},
has prevented evaluation with larger collections. However, more efficient fully \tblEmb{}-based and \tblCoop{} methods now enable evaluation on larger collections and in realistic scenarios.


We thus increase the benchmark size by merging test instances from different available evaluation sets. In particular, we construct a collection spanning 20k images: it blends the test sets of MSCOCO (5k instances), Flickr30k (1k), and the development set of CC (14k). 
Note that we simply augment the benchmarks but the query set with labels for each standardized evaluation task/set remains unchanged; in other words, the instances from other datasets are used as distractors that increase the search space and make the retrieval task more difficult. The results thus provide insights into the model performance in the target domain, as well as its robustness regarding out-of-distribution data.  We now observe in Table~\ref{tab:res:enlarge-benchmark} more salient performance differences, which were lacking with the smaller benchmarks. The pure \tblEmb{}-based approach now substantially underperforms \textsc{Sep/Joint+Coop} variants. The \textsc{Joint+Coop} does remain the best-scoring variant overall.

\begin{table}[]
    \footnotesize
    \centering
    \def\arraystretch{0.85}
    \resizebox{0.99\columnwidth}{!}{
    \begin{tabular}{l rrr rrr}
    \toprule
         \bf Model & \multicolumn{3}{c}{\bf Image Retrieval} & \multicolumn{3}{c}{\bf Text Retrieval} \\
         & R@1 & R@5 & R@10 & R@1 & R@5 & R@10 \\
\cmidrule(lr){2-4} \cmidrule(lr){5-7}
\rowcolor{Gray}
& \multicolumn{6}{c}{\underline{Flickr30k 1k}
+ \textit{CC 14k} + \textit{MSCOCO 5k}} \\
\cmidrule(lr){2-4} \cmidrule(lr){5-7}
\tblEmb{}$^\text{OSCAR}$ & 45.8 & 69.1 & 76.1 & 71.1 & 90.9 & 94.9 \\
\textsc{Sep+Coop}$^\text{OSCAR}$ & 55.5 & 75.8 & 80.1 & 80.5 & \textbf{93.8} & \textbf{95.4} \\
\textsc{Joint+Coop}$^\text{OSCAR}$ & \textbf{55.9} & \textbf{77.5} & \textbf{82.9} & \textbf{81.0} & 92.9 & 94.9 \\

\cmidrule(lr){2-4} \cmidrule(lr){5-7}
& \multicolumn{6}{c}{\underline{MSCOCO 5k}
+ \textit{CC 14k} + \textit{Flickr 1k}} \\
\cmidrule(lr){2-4} \cmidrule(lr){5-7}
\tblEmb{}$^\text{OSCAR}$ & 40.6 & 68.5 & 78.1 & 62.5 & 87.7 & 93.3 \\
\textsc{Sep+Coop}$^\text{OSCAR}$ & 43.7 & 72.1 & 81.2 & 68.2 & \textbf{90.4} & 94.3 \\
\textsc{Joint+Coop}$^\text{OSCAR}$ & \textbf{45.6} & \textbf{73.0} & \textbf{82.3} & \textbf{69.0 }& 90.3 & \textbf{94.7} \\
        \bottomrule
    \end{tabular}
    }
    \caption{Results with larger benchmarks. The dataset \underline{underlined} 
    indicates the actual standard task with the corresponding task data and labels used, while the instances from the datasets in \textit{italic} are used as additional non-relevant test examples (i.e., distractors in the search space).  
    }
    \label{tab:res:enlarge-benchmark}
\end{table}

\begin{table*}[]
    \footnotesize
    \centering
        \def\arraystretch{0.93}

\resizebox{\textwidth}{!}{
    \begin{tabular}{l|rrr|rrr|rrr|rrr|rrr|rrr}
    \toprule
         \bf Loss &  \multicolumn{3}{c}{\bf Image Retrieval} & \multicolumn{3}{c}{\bf Text Retrieval}  & \multicolumn{3}{|c}{\bf Image Retrieval} & \multicolumn{3}{c}{\bf Text Retrieval} & \multicolumn{3}{|c}{\bf Image Retrieval} & \multicolumn{3}{c}{\bf Text Retrieval}\\
          & R@1 & R@5 & R@10 & R@1 & R@5 & R@10 & R@1 & R@5 & R@10 & R@1 & R@5 & R@10 & R@1 & R@5 & R@10 & R@1 & R@5 & R@10\\
\cmidrule(lr){2-4} \cmidrule(lr){5-7} \cmidrule(lr){8-10} \cmidrule(lr){11-13} \cmidrule(lr){14-16} \cmidrule(lr){17-19}
& \multicolumn{6}{c|}{MSCOCO 5k} & \multicolumn{6}{c|}{Flickr30k 1k}  & \multicolumn{6}{c}{CC 14k} \\
\cmidrule(lr){2-4} \cmidrule(lr){5-7} \cmidrule(lr){8-10} \cmidrule(lr){11-13} \cmidrule(lr){14-16} \cmidrule(lr){17-19}
 \tblConj{}$_\text{In-Domain}^\text{OSCAR}$ &  \it  54.7 & \it 81.3 &\it  88.9 & \it 70.8 & \it 91.0 & \it 95.2  & \it 76.4 & \it 93.6 & \it  96.2 & \it  89.4 & \it 97.7 & \it 99.0  & \multicolumn{1}{c}{\, ---} & \multicolumn{1}{c}{\, ---} & \multicolumn{1}{c|}{\, ---} & \multicolumn{1}{c}{\, ---} & \multicolumn{1}{c}{\, ---} & \multicolumn{1}{c}{\, ---} \\
\cmidrule(lr){2-4} \cmidrule(lr){5-7} \cmidrule(lr){8-10} \cmidrule(lr){11-13} \cmidrule(lr){14-16} \cmidrule(lr){17-19} 
\tblCross{}$^\text{UNITER}$ &  \multicolumn{1}{c}{\, ---} & \multicolumn{1}{c}{\, ---} & \multicolumn{1}{c|}{\, ---} & \multicolumn{1}{c}{\, ---} & \multicolumn{1}{c}{\, ---} & \multicolumn{1}{c|}{\, ---} & 66.2 & 88.4 & 92.9 & 80.7 & 95.7 & 98.0  & \multicolumn{1}{c}{\, ---} & \multicolumn{1}{c}{\, ---} & \multicolumn{1}{c|}{\, ---} & \multicolumn{1}{c}{\, ---} & \multicolumn{1}{c}{\, ---} & \multicolumn{1}{c}{\, ---} \\
\tblCross{}$^\text{OSCAR}$ &  \bf 47.8 & \bf 75.7 & \bf 84.6 & 61.8 & \bf 86.2 & \bf 92.0 & 67.2 & 88.5 & 92.7 & 81.0 & 95.5 & 97.8  & \multicolumn{1}{c}{\, ---} & \multicolumn{1}{c}{\, ---} & \multicolumn{1}{c|}{\, ---} & \multicolumn{1}{c}{\, ---} & \multicolumn{1}{c}{\, ---} & \multicolumn{1}{c}{\, ---} \\
\cmidrule(lr){2-4} \cmidrule(lr){5-7} \cmidrule(lr){8-10} \cmidrule(lr){11-13} \cmidrule(lr){14-16} \cmidrule(lr){17-19}
CLIP & 30.4 & 56.1 & 66.9 & 50.1 & 74.8 & 83.6 & 61.1 & 85.9 & 91.8 & 81.9 & 95.0 & 97.5 &\bf 30.8 & \bf52.7 & \bf61.3 & \bf32.1 & \bf53.9 &\bf 63.0\\
\tblEmb{}$^\text{OSCAR}$ &  37.6 & 64.4 & 75.0 & 52.0 & 78.1 & 86.3 & 63.3 & 86.4 & 91.6 & 78.2 & 94.0 & 97.3 & 13.8 & 29.4 & 37.9 &  14.4 & 29.6 & 37.6\\
\tblCoop{}$^\text{OSCAR}$ &  47.6 & 73.9 & 81.2 & 62.8 & 83.8 & 88.7  & 67.6 & 89.0 & 93.1 & 82.4 & 96.3 & \bf 98.2 & 16.8 & 34.3 & 41.9 & 17.0 & 33.5 & 41.5\\
\tblConj{}$^\text{OSCAR}$ &   47.6 & 74.5 &  82.6 & \bf 63.9 & 85.7 & 91.0 & \bf 70.0 & \bf 90.2 & \bf 94.1 & \bf 83.7 & \bf 96.8 & 97.9 & 16.7 & 34.7 & 43.6 & 17.5 & 34.6 & 43.5 \\

        \bottomrule
    \end{tabular}
    }
    \caption{Results for zero-shot evaluation on Flickr30k, MSCOCO, and CC. For Flickr30k and MSCOCO  results we train on the respective other datasets. For CC results we train on Flickr30k. \tblConj{}$_\text{In-Domain}^\text{OSCAR}$ is the in-domain performance for the \tblConj{} approach and here represents the upper-bound.
    }
    \label{tab:res:zs}
\end{table*}

\begin{table}[ht]
    \footnotesize
    \centering
    \def\arraystretch{0.93}
    \resizebox{0.99\columnwidth}{!}{
    \begin{tabularx}{0.98\linewidth}{l|X:XXX|X}
    \toprule
         \bf Model & \bf en & \bf de & \bf fr & \bf cs & \bf  Avg \\
         \cmidrule(lr){2-2} \cmidrule(lr){3-5} \cmidrule(lr){6-6}
\tblCross{}$^\text{M3P}$  \cite{Huang2020} & 86.0 & 48.8 & 39.4 & 38.8 & \it 42.3\\
\cmidrule(lr){2-2} \cmidrule(lr){3-5} \cmidrule(lr){6-6}
\tblEmb{}$^\text{M3P}$ & 81.3 & 52.4 & 49.7 & 39.6 & \it 47.2\\
\tblCross{}$^\text{M3P}$  & 84.2 & 52.6 & 49.6 & 33.4 & \it 45.2\\
\tblCoop{}$^\text{M3P}$ & 84.4 & \bf 55.6 & \bf  52.2 & \bf  39.8 & \textit{\textbf{49.2}}\\
\tblConj{}$^\text{M3P}$ & 83.5 & 54.2 & 48.4 & 39.4 & \it 47.3\\
        \bottomrule
    \end{tabularx}
    }
    \caption{Multilingual image-text retrieval results (in mR) on Multi30k. Models are trained on the English data. Avg results of non-English languages.}
    \label{tab:res:multilingual-en}
\end{table}

\vspace{1.6mm}
\noindent \textbf{Zero-Shot Performance.}
Relying on multi-modal and multilingual representations fine-tuned for cross-modal retrieval, the proposed methods should also generalize to new unseen captions and images beyond the dataset used for fine-tuning. Therefore, we directly transfer the model fine-tuned on one dataset to the test data of another dataset (e.g., fine-tune on MSCOCO data, test on Flickr30k). As baselines, we use the reported zero-shot results of UNITER \cite{Chen2019} for Flickr30k\footnote{They do not report results for MSCOCO.} and we also evaluate the CLIP model.\footnote{
We use the ViT-B/32 model variant. Retrieval results from \citet{radford2021learning} Table 13 use the (larger) ViT-L/14 variant that has not been released to the public.
}

The zero-shot results in Table~\ref{tab:res:zs}, reveal that the CE variant slightly outperforms other approaches when transferring from Flickr30k to MSCOCO, while \textsc{Joint+Coop}$^\text{OSCAR}$ remains competitive. However, for the opposite direction, we achieve considerable performance gains with the \textsc{Joint+Coop}$^\text{OSCAR}$ variant. On CC, all variants considerably underperform CLIP; we speculate that it might be due to a more diverse set of images included in CC, including illustrations, which neither exist in MSCOCO nor Flickr30k. This means that CLIP has a considerable advantage on CC due to its exposure to massive amounts of data during pretraining. 

Multilingual zero-shot results, where we fine-tune on the English Multi30k captions and test on the captions in other languages, are shown in Table~\ref{tab:res:multilingual-en}. Cooperative approaches again excel; the highest scores are achieved by \textsc{Sep+Coop}$^\text{M3P}$.

\begin{figure}[!t]
    \centering
    \includegraphics[width=0.99\linewidth]{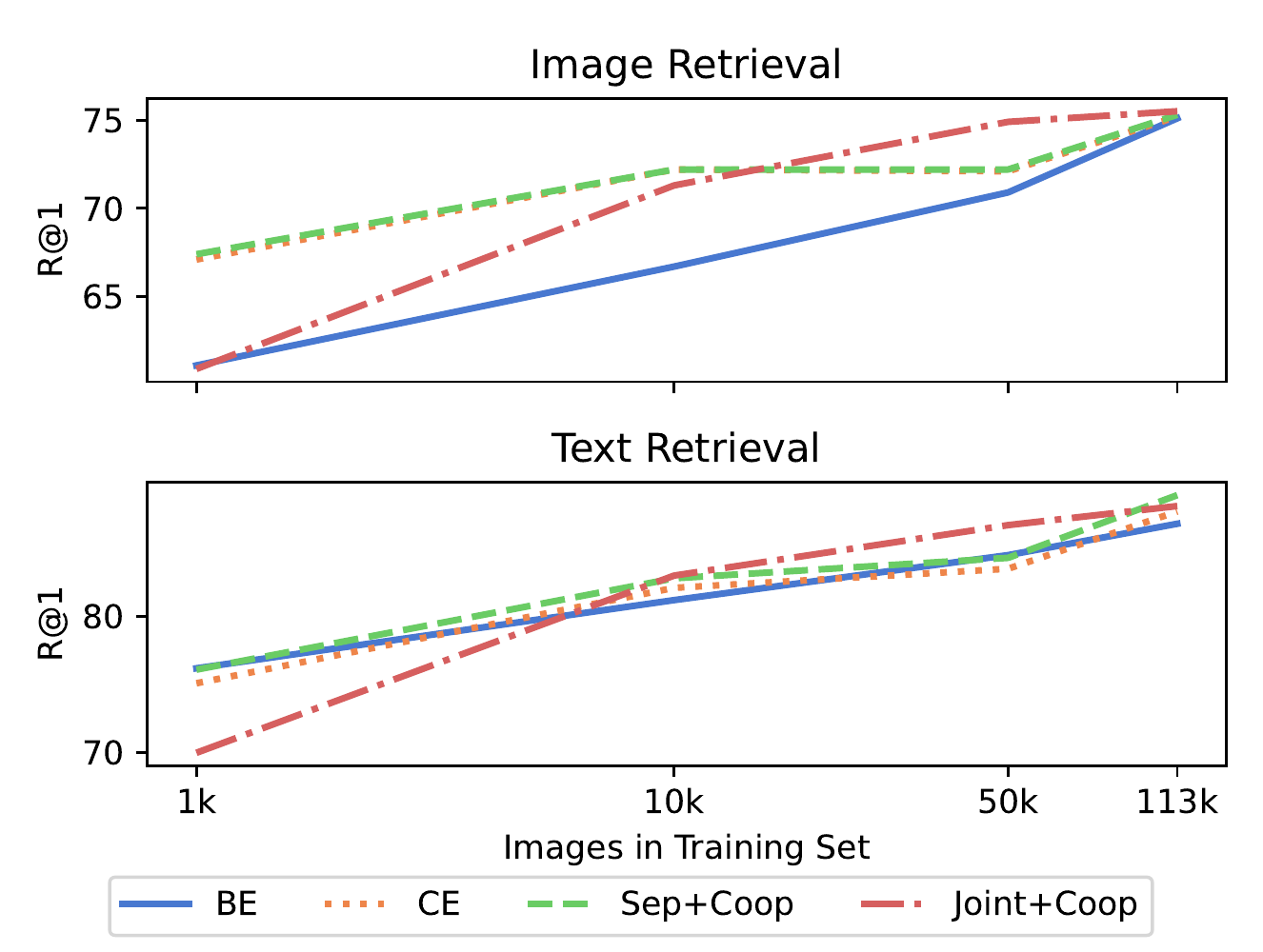}
    \caption{Impact of data size for fine-tuning on retrieval performance. MSCOCO training and test data; OSCAR as the underlying Transformer.  
    }
    \label{fig:res:small-train}
\end{figure}

\vspace{1.6mm}
\noindent \textbf{Sample Efficiency.}
We also analyze how the amount of image-text data for fine-tuning impacts the retrieval performance; we thus sample smaller datasets from the full MSCOCO training set, covering 1k, 10k, and 50k images with their captions (5 per image). The results in Figure~\ref{fig:res:small-train} reveal that \tblEmb{}-based approaches in general are considerably less sample-efficient than cross-encoders. They particularly struggle in the lowest-data scenario with only 1k images available; this is also reflected in the lower performance of \textsc{Joint+Coop} in the 1k setup. A reason behind the more effective adaptation of \tblCross{} to low-data regimes might be their richer ``input consumption'': with 1k images and 5k captions, \tblCross{} runs a whole grid of 1k$\times$5k items through its network, which provides more learning signal with fewer data available. On the other hand, \tblEmb{}-based approaches are expected to learn effective encoders of both modalities separately based solely on 1k images and 5k captions, without any cross-modal interaction.

\begin{table*}[ht]
    \footnotesize
    \centering
\resizebox{\textwidth}{!}{
    \begin{tabular}{lr|rrr|rrr|rrr|rrr|rrr|rrr}
    \toprule
         \bf Model &  $\mathbf{k}$ & \multicolumn{3}{c}{\bf Image Retrieval} & \multicolumn{3}{c}{\bf Text Retrieval} & \multicolumn{3}{|c}{\bf Image Retrieval} & \multicolumn{3}{c}{\bf Text Retrieval} & \multicolumn{3}{|c}{\bf Image Retrieval} & \multicolumn{3}{c}{\bf Text Retrieval}\\
         && R@1 & R@5 & R@10 & R@1 & R@5 & R@10 & R@1 & R@5 & R@10 & R@1 & R@5 & R@10 & R@1 & R@5 & R@10 & R@1 & R@5 & R@10 \\
\cmidrule(lr){3-5} \cmidrule(lr){6-8} \cmidrule(lr){9-11} \cmidrule(lr){12-14} \cmidrule(lr){15-17} \cmidrule(lr){18-20}
&& \multicolumn{6}{c}{MSCOCO 1k} & \multicolumn{6}{|c|}{MSCOCO 5k} & \multicolumn{6}{c}{Flickr30k}\\
\cmidrule(lr){3-5} \cmidrule(lr){6-8} \cmidrule(lr){9-11} \cmidrule(lr){12-14} \cmidrule(lr){15-17} \cmidrule(lr){18-20}
\multirow{3}{*}{\tblCoop{}}  & 10 & \bf 75.4 & 94.8 & 97.2 & \bf 88.4 & 98.8 & 99.7& \bf 53.2 & 80.3 & 86.6 & \bf 71.1 & 90.9 & 94.3 & 75.9 & 92.2 & 93.4 & \bf 89.2 & 97.8 & 98.4 \\
& 20 & 75.3 & \bf 95.2 & 98.1 & 87.9 & 98.9 & \bf 99.8& 52.8 & \bf 80.5 & \bf 88.5 & 70.2 & \bf 91.6 & 95.0 & \bf 76.0 & 93.0 & 95.0 & 88.7 & 98.3 & 99.2 \\
 &50 & 75.2 & 95.0 & \bf 98.2 & 87.9 & \bf 99.1 & \bf 99.8& 52.6 & 80.1 & 88.4 & 70.1 & 91.4 & \bf 95.5 & 75.9 & \bf 93.4 & \bf 96.3 & 88.9 & \bf 98.4 & \bf 99.4\\
\cmidrule(lr){3-5} \cmidrule(lr){6-8} \cmidrule(lr){9-11} \cmidrule(lr){12-14} \cmidrule(lr){15-17} \cmidrule(lr){18-20}
\multirow{3}{*}{\tblConj{}}  &10 & 75.4 & \bf 95.5 & 97.8 & 88.0 & \bf 98.8 & \bf 99.9& \bf 54.8 & 81.2 & 88.0 & \bf 70.9 & \bf 91.2 & 95.0 & \bf 76.5 & 93.2 & 95.0 & 88.9 & 97.3 & 98.6 \\
 &20 & \bf 75.5 & 95.4 & 98.2 & 88.1 & 98.6 & 99.5& 54.7 & \bf 81.3 & \bf  88.9 & 70.8 & 91.0 & 95.2 & 76.4 & \bf 93.6 & 96.2 & \bf 89.4 & 97.7 & \bf  99.0\\
& 50 & 75.4 & 95.4 & \bf 98.3 & \bf 88.2 & 98.4 & 99.4& 54.6 & 81.2 & 88.8 & 70.7 & 91.1 & \bf 95.3 & \bf 76.5 & 93.5 & \bf 96.5 & 89.1 & \bf 98.0 & 98.9\\

        \bottomrule
    \end{tabular}
    }
    \caption{ 
    Results with $\tblCoop{}$ and $\tblConj{}$ reranking the top-$k$ candidates. \textbf{Bold} numbers indicate which $k$ value resulted in the highest score for each separate model.}
    \label{tab:res:rr-k}
\end{table*}

\begin{table}[ht!]
    \footnotesize
    \centering
    \resizebox{\linewidth}{!}{
    \begin{tabular}{llr|rrr|rrr}
    \toprule
         \bf Model & \bf Sum & $\lambda$ &\multicolumn{3}{c}{\bf Image Retrieval} & \multicolumn{3}{c}{\bf Text Retrieval} \\
         &&& R@1 & R@5 & R@10 & R@1 & R@5 & R@10 \\
\cmidrule(lr){2-3} \cmidrule(lr){4-6} \cmidrule(lr){7-9}
\multirow{7}{*}{\tblCoop{}}  & - &  & 76.0 & 93.0 & 95.0 & 88.7 & 98.3 & 99.2 \\
\cmidrule(lr){2-3} \cmidrule(lr){4-6} \cmidrule(lr){7-9}
  &\multirow{3}{*}{\textsc{add}} &0.1 & 76.0 & 92.7 & 94.8 & 86.4 & 98.7 & 99.2 \\
  &  &0.5 & 75.7 & 92.6 & 94.7 & 85.9 & 98.5 & 99.2 \\
  &  &0.9 & 74.5 & 92.5 & 94.7 & 85.1 & 98.3 & 99.2 \\
\cmidrule(lr){2-3} \cmidrule(lr){4-6} \cmidrule(lr){7-9}
  & \multirow{3}{*}{\textsc{norm\_add}} &0.1 & 70.8 & 90.2 & 93.8 & 86.2 & 98.5 & 99.2 \\
  & &0.5 & 70.7 & 90.3 & 93.7 & 85.4 & 98.4 & 99.2 \\
  & &0.9 & 70.3 & 90.1 & 93.7 & 83.8 & 97.6 & 98.8 \\
\cmidrule(lr){2-9}
\multirow{7}{*}{\tblConj{}}  & - &  & 76.4 & 93.6 & 96.2 & 89.4 & 97.7 & 99.0 \\
\cmidrule(lr){2-3} \cmidrule(lr){4-6} \cmidrule(lr){7-9}
 &\multirow{3}{*}{\textsc{add}} &0.1 & 76.7 & 93.3 & 95.8 & 88.5 & 98.0 & 99.1 \\
 & &0.5 & 75.6 & 93.1 & 95.5 & 87.2 & 97.8 & 99.1 \\
 & &0.9 & 74.6 & 92.8 & 95.5 & 87.3 & 97.8 & 99.1 \\
\cmidrule(lr){2-3} \cmidrule(lr){4-6} \cmidrule(lr){7-9}
 & \multirow{3}{*}{\textsc{norm\_add}}  &0.1 & 72.8 & 92.0 & 95.2 & 87.6 & 97.9 & 99.2 \\
 &  &0.5 & 72.5 & 92.0 & 95.2 & 87.3 & 97.9 & 99.0 \\
 &  &0.9 & 72.3 & 91.8 & 95.2 & 86.4 & 97.0 & 99.0 \\
        \bottomrule
    \end{tabular}
    }
    \caption{Results on Flickr30k for different combinations of the embedding and cross-encoder scores using the  functions $\textsc{add}_\lambda$ and $\textsc{norm\_add}_\lambda$ and different values for $\lambda$. - indicates the results for reranking using only the cross-encoder.}
    \label{tab:res:rr-sum}
\end{table}

\begin{figure}[!t]
    \centering
\includegraphics[trim={0 0 0 8mm}, clip, width=0.8\linewidth]{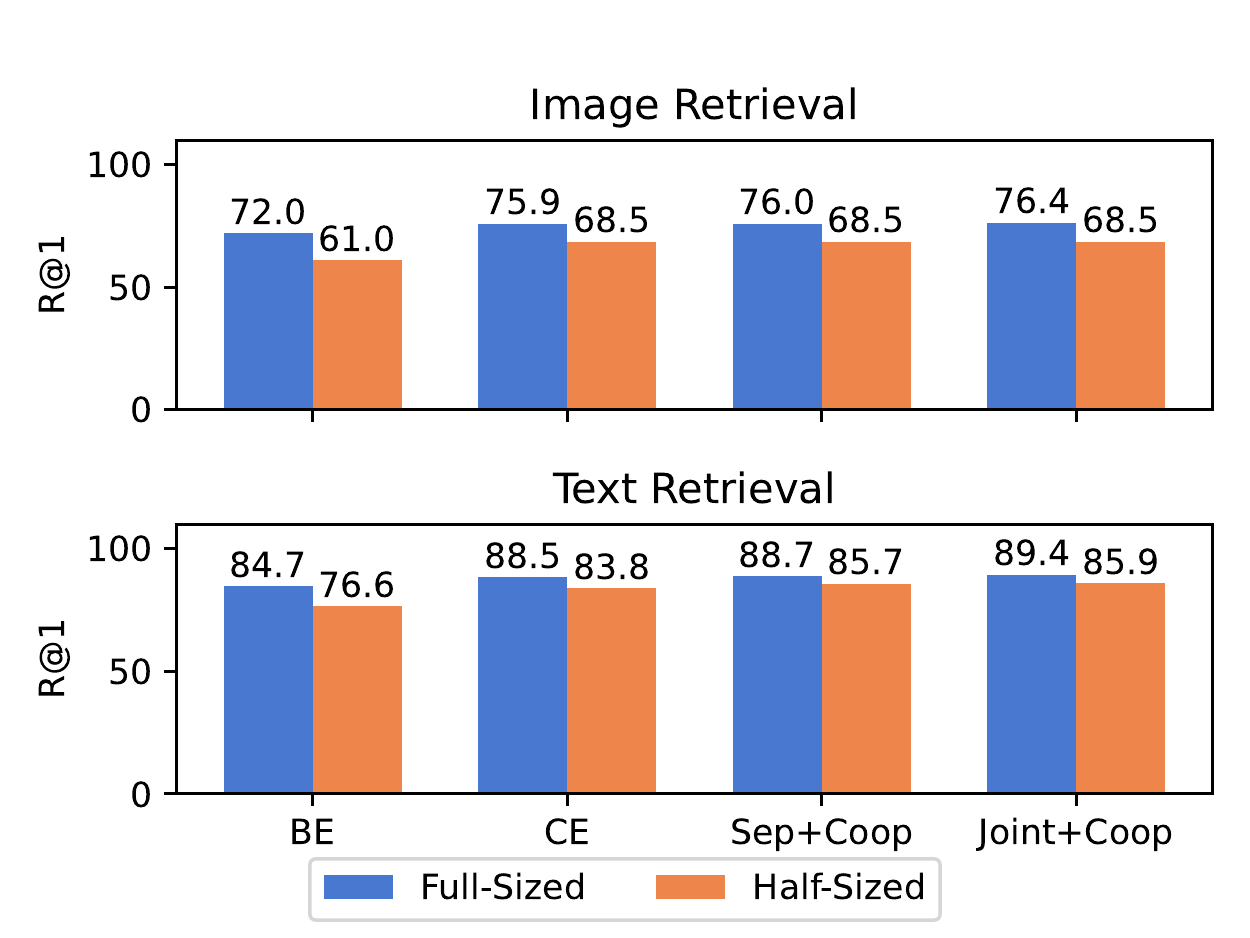}
    \caption{Half- vs. full-sized models on Flickr30k. With half-sized models, we skip every odd-numbered Transformer layer.}
    \label{fig:half-sizes_r1}
\end{figure}

\vspace{1.4mm}
\noindent \textbf{Parameter Efficiency.}
We also provide a simple parameter efficiency analysis by initializing the models with pretrained OSCAR weights, but only passing the representations through every second layer, effectively halving the total amount of Transformer parameters. The results are shown in Figure~\ref{fig:half-sizes_r1}. 
The performance with all approaches using the ``halved'' model is around $\sim 90\%$ of the performance with the full Transformer. Overall, the \textsc{Joint+Coop} method again achieves the highest scores. This suggests that the proposed fine-tuning approaches are applicable also to smaller models, with similar relative trends in retrieval results.

\vspace{1.6mm}
\noindent \textbf{Retrieving Top-$k$. }
We analyze different values for $k$ for top-$k$ retrieval of the BE component in Table \ref{tab:res:rr-k}. Selecting small values for $k$ significantly decreases the retrieval latency, as less instances need to be cross-encoded. However, selecting $k$ values that are too small can come at a cost of precision, as the true positive instance might not be among the top-$k$ retrieved instances of the \tblEmb{} model.  
In our experiments, $k=20$ achieves the best trade-off between precision and retrieval latency.

\vspace{1.6mm}
\noindent\textbf{Combining Ranking.}
We evaluate the ranking score combination of the two components \textsc{Joint+BE} and \textsc{Joint+CE} in Table~\ref{tab:res:rr-sum}. 
We combine the ranking of the bi-encoder submodel and the cross-encoder submodel by summing over the scores using two different variations:

(1) We directly add the scores  in a weighted sum   
\begin{equation}
\textsc{add}_\lambda(e, c) = \lambda e + (1-\lambda) c
\end{equation} 
where  $e$ and  $c$ are the embedding cosine similarity and cross-encoder similarity scores respectively and  $\lambda$ is a weighting parameter.
The cross-encoder scores have been processed with a sigmoid function so that both $e$ and $c$ are in the same value range. 
The final ranking is then defined by $\textsc{add}_\lambda(e, c)$. 

(2) We  
separately 0-1-normalize the scores for the top-$k$ candidates of the bi- and cross-encoder
before combining them for $\textsc{norm\_add}_\lambda(e, c)$, which is defined analog to $\textsc{add}_\lambda(e, c)$.  

However, we find that relying solely on the cross-encoder achieves the best results.
This suggests that the scores by the bi-encoder are useful in the ``global'' scope with all data to retrieve strong candidates but in the ``local'' scope of the top-$k$ candidates, the cross-encoder is superior.

\section{Conclusion}

We proposed a novel framework that converts pretrained multi-modal Transformers into effective \textit{and} efficient cross-modal retrieval models. The framework is applicable to any pretrained model and combines the efficiency of bi-encoder (\tblEmb{}) approaches with the accuracy of computationally more demanding cross-encoding (\tblCross{}) approaches. Their synergistic effect at retrieval is achieved through a cooperative retrieve-and-rerank regime, where the initial retrieval from a large collection is performed via efficient \tblEmb{} approaches, followed by another accuracy-driven step via a \tblCross{} model. Moreover, we introduced a parameter-efficient joint fine-tuning regime that blends \tblEmb{} and \tblCross{} into a single model with shared weights. Our results with state-of-the-art pretrained models across a range of standard monolingual and multilingual cross-modal retrieval tasks and setups validated the strong performance of such cooperative and joint approaches. At the same time, we demonstrated their retrieval efficiency, which makes them viable in realistic retrieval scenarios with large  collections. In future work, we will put more focus on zero-shot and few-shot retrieval scenarios, and expand the approach to more languages, modalities, and tasks.

\section*{Acknowledgments}

Ubiquitous Knowledge Processing Lab acknowledge the financial support of the German Federal Ministry of Education and Research (BMBF) under the promotional reference 13N15897 (MISRIK), the LOEWE initiative (Hesse, Germany) within the emergenCITY center, and the German Research Foundation (DFG) as part of the UKP-SQuARE project (grant GU 798/29-1).
The work of \textit{Ivan Vuli\'{c}} has been supported by the ERC Consolidator Grant LEXICAL (no 648909), ERC PoC Grant MultiConvAI (no. 957356), and a research donation from Huawei. 

We thank Kevin Stowe and Christopher Klamm for insightful feedback and suggestions on a draft of this paper and we thank the TACL reviewers and Action Editor for their valuable feedback and comments during the editing process.

\bibliography{anthology,acl2021}
\bibliographystyle{acl_natbib}

\end{document}